\documentclass{article}

\usepackage{arxiv}

\usepackage[utf8]{inputenc} 
\usepackage[T1]{fontenc}    
\usepackage{hyperref}       
\usepackage{url}            
\usepackage{booktabs}       
\usepackage{amsfonts}       
\usepackage{nicefrac}       
\usepackage{microtype}      
\usepackage{lipsum}		
\usepackage{graphicx}
\usepackage{natbib}
\usepackage{doi}
\usepackage{subcaption}

\usepackage{amssymb}
\usepackage{mathtools}
\DeclarePairedDelimiter{\norm}{\lVert}{\rVert}
\DeclareMathOperator*{\argmin}{arg\,min}

\usepackage{amsthm}

\newcommand*{\transpose}{{\mkern-1.5mu\mathsf{T}}}

\usepackage{algorithm}
\usepackage{algorithmic}

\title{Incorporating Fairness Constraints into Archetypal Analysis}

\date{} 					

\author{Aleix Alcacer \\
	Departament de Matemàtiques\\
	Universitat Jaume I, Spain\\
	\texttt{aalcacer@uji.es} \\
	\And
	Irene Epifanio \\
	Departament de Matemàtiques\\
	Universitat Jaume I, Spain\\
	\texttt{epifanio@uji.es} \\
}

\begin{document}
\maketitle

\begin{abstract}
Archetypal Analysis (AA) is an unsupervised learning method that represents data as convex combinations of extreme patterns called archetypes. While AA provides interpretable and low-dimensional representations, it can inadvertently encode sensitive attributes, leading to fairness concerns. In this work, we propose Fair Archetypal Analysis (FairAA), a modified formulation that explicitly reduces the influence of sensitive group information in the learned projections. We also introduce FairKernelAA, a nonlinear extension that addresses fairness in more complex data distributions. Our approach incorporates a fairness regularization term while preserving the structure and interpretability of the archetypes. We evaluate FairAA and FairKernelAA on synthetic datasets, including linear, nonlinear, and multi-group scenarios, demonstrating their ability to reduce group separability—as measured by mean maximum discrepancy and linear separability—without substantially compromising explained variance. We further validate our methods on the real-world ANSUR I dataset, confirming their robustness and practical utility. The results show that FairAA achieves a favorable trade-off between utility and fairness, making it a promising tool for responsible representation learning in sensitive applications.
\end{abstract}


\section{Introduction}
Recent advancements in machine learning (ML) have significantly enhanced the ability of computational systems to reason within complex domains. Although ML has shown success in decision-making processes, there is increasing concern about the potential for discriminatory biases in its decision rules. Specially when ML is integrated into critical systems such as finance \cite{tsai2010credit}, hiring \cite{raghavan2020mitigating}, education \cite{waters2014grade}, healthcare \cite{shehab2022machine}, and the legal domain \cite{ghasemi2021application}, it is essential to design fair and accurate algorithms that avoid bias in sensitive features like race or gender.

Despite the growing efforts to address fairness in supervised learning, the issue of fairness in unsupervised learning tasks remains relatively underexplored \cite{chhabra2021overview}. Note that the first work on fair clustering was published in 2017 \cite{chierichetti2017fair}. Ensuring fairness in unsupervised ML setting is more challenging than in supervised ML tasks because the data lacks labels, making it impossible to compute ground-truth error rates to assess bias and unfairness. As a result, both defining and enforcing fairness in unsupervised ML tools present significant difficulties. However, ensuring fairness in unsupervised learning models is just as important as in supervised learning models since unsupervised methods, such as dimensionality reduction and clustering, are widely used for tasks like data visualization, uncovering common patterns or trends, and reducing data size, among other applications. Their impact could be critical in decision-making systems, especially if they treat different groups in the data unfairly. Note also that unsupervised learning tasks, such as dimensionality reduction, often precede supervised learning tasks.

Recently, methods have been developed to integrate fairness into unsupervised techniques like clustering and principal component analysis (PCA). Despite these advancements, fairness remains unaddressed in other unsupervised methods. This is the case of archetypal analysis (AA), an unsupervised technique that lies halfway between clustering and PCA \cite{morup2012archetypal}. Data decomposition techniques are commonly employed to uncover latent components within datasets, representing the data as a linear combination of several underlying factors. The specific nature of these techniques depends on the constraints imposed on the factors and the way they are combined, leading to different unsupervised methods with distinct objectives \cite{vinue2015archetypoids}. At one end of the spectrum lies PCA, which effectively captures data variability but often sacrifices interpretability, as its components are constructed as linear combinations of all input variables. At the opposite end are clustering techniques such as k-means, which offer highly interpretable factors—cluster centroids in the case of k-means. However, these methods lack modeling flexibility due to their binary assignment of data points to clusters. AA occupies a middle ground between these approaches. It provides factors that are as interpretable as those from clustering methods while offering greater modeling flexibility. \cite{alcacer2025survey} present a table that summarizes the relationships among various unsupervised multivariate techniques.

As regards fairness in clustering, there are numerous fairness definitions for clustering, with different studies adopting various metrics or introducing new ones. Additionally, the methods used to enforce fairness constraints differ significantly in their approaches and underlying methodologies. \cite{chhabra2021overview} have categorized clustering fairness notions into four groups: group-level, individual-level, algorithm agnostic, and algorithm specific fairness. 

As regards fairness in PCA, there are also different approaches. One of them consists of identifying the optimal projection in which no linear classifier can accurately infer demographic information from the transformed data \cite{kleindessner2023efficient,olfat2019convex,lee2022fast,d45fa562190149b3bbd83b502a9b08ca}. A fundamentally different notion of fair PCA consists of balancing the excess reconstruction error among different demographic groups \cite{samadi2018price,kamani2022efficient}.
Besides the previous approaches for fair PCA, other fair representation learning have been proposed, such as based on variational autoencoder \cite{DBLP:journals/corr/LouizosSLWZ15} or formulating the problem as an adversarial game \cite{DBLP:journals/corr/EdwardsS15,madras2018learning}.

Furthermore, fairness-aware approaches can be broadly categorized based on the stage of the learning pipeline at which fairness is incorporated. During preprocessing, i.e. before learning the model; in-processing, i.e. while the model is being learned; and post-processing, after the model has been learned. Most methods use the in-processing approach.

In this work, we propose to introduce fairness constraints into AA for the first time. An in-processing approach for enforcing fairness constraints on AA space is introduced based on the idea of obtaining a fair representation learning \cite{zemel2013learning}.
The goal is to learn a data representation that conceals demographic information while retaining as much relevant (non-demographic) information as possible. To this end, we introduce a suitable form of regularization. Similar forms of regularization have previously been applied in the context of AA to address other problems, such as enforcing sparsity in the representations \cite{xu2022l,xu2023manifold}

 The outline of the paper is as follows: in Sect. \ref{sec:fairAA}  we introduce our proposal for incorporating fairness constraints into AA. In Sect. \ref{sec:extensions}, we explore various extensions of our proposal. In Sect. \ref{sec:res} we apply our proposal to several artificial and real data sets and compare it to AA. Finally, in Sect. \ref{sec:con} we present our conclusions and highlight potential directions for future research.


\section{Fair archetypal analysis}
\label{sec:fairAA}

\subsection{Archetypal analysis}

Originally introduced in \cite{cutler1994archetypal}, AA is a data representation technique that models each data point as a convex combination of a set of extremal archetypes.

In this framework, the data matrix $\mathbf{X} \in \mathbb{R}^{n \times d}$ consists of $n$ data points in a $d$-dimensional space, while the coefficient matrices $\mathbf{S} \in \mathbb{R}^{n \times k}$ and $\mathbf{C} \in \mathbb{R}^{k \times n}$ enable a two-step decomposition: each data point is expressed as a convex combination of $k$ archetypes, which, in turn, are convex combinations of the original data points.

Formally, AA is defined as
$$\argmin_{\mathbf{S}, \mathbf{C}} \norm{\mathbf{X} - \mathbf{SCX}}_F^2 $$
subject to $\norm{\mathbf{s}_n}_2 = 1$, $\mathbf{s}_n \geq 0$, $\norm{\mathbf{c}_k}_2 = 1$ and $\mathbf{c}_k \geq 0$.

This formulation ensures that each data point is represented as a convex combination of archetypes, which themselves are convex combinations of the original data points.

As proposed in \cite{morup2012archetypal}, the objective function can be rewritten in terms of the trace operator as
$$E = \norm{\mathbf{X} - \mathbf{SCX}}_F^2 = \text{tr} \left( \mathbf{X} \mathbf{X}^\transpose - 2 \mathbf{SCX}\mathbf{X}^\transpose +  \mathbf{SCX}\mathbf{X}^\transpose \mathbf{C}^\transpose \mathbf{S}^\transpose \right),$$
which allows us to derive the gradients with respect to $\mathbf{S}$ and $\mathbf{C}$
$$\nabla_{\mathbf{S}} E = 2(\mathbf{SCX} \mathbf{X}^\transpose\mathbf{C}^\transpose -\mathbf{X} \mathbf{X}^\transpose\mathbf{C}^\transpose),$$
$$\nabla_{\mathbf{C}} E = 2(\mathbf{S}^\transpose\mathbf{S}\mathbf{C}\mathbf{X}\mathbf{X}^\transpose - \mathbf{S}^\transpose\mathbf{X}\mathbf{X}^\transpose).$$

Thus, to minimize $E$, AA can be computed by alternately updating
$$\mathbf{S} \leftarrow \mathbf{S} - \eta_S \cdot \nabla_{\mathbf{S}} E,$$
$$\mathbf{C} \leftarrow \mathbf{C} - \eta_C \cdot \nabla_{\mathbf{C}} E,$$
where $\eta_S$ and $\eta_C$ are step size parameters controlling the rate of convergence. However, since these gradient updates may move $\mathbf{S}$ and $\mathbf{C}$ outside of their respective constraint sets, a projection step is required. Specifically, the rows of $\mathbf{S}$ and $\mathbf{C}$ must be projected back onto their feasible regions, which correspond to the standard $K$- and $N$-simplex, respectively. This ensures that the convexity constraints remain satisfied throughout the optimization process, preserving the interpretability of archetypes as extremal points in the data space.

\subsection{Formulation of fair  archetypal analysis}

Following \cite{kleindessner2023efficient} on fair PCA, our objective in fair archetypal analysis (FairAA) is to obscure critical information when projecting the dataset onto the archetypal space.

Specifically, we aim for a projection that eliminates critical information. Let $z_i \in \{0, 1\}$ represent a critical attribute of the data point $x_i$, indicating membership in one of two groups. Ideally, no classifier should be able to infer $z_i$ from the projection of $x_i$ onto the $k$-archetypal space. Formally, FairAA seeks to solve the following optimization problem:

$$\argmin_{\mathbf{S}, \mathbf{C}} \norm{\mathbf{X} - \mathbf{SCX}}_F^2 $$

subject to $\norm{\mathbf{s}_n}_2 = 1$, $\mathbf{s}_n \geq 0$, $\norm{\mathbf{c}_k}_2 = 1$, $\mathbf{c}_k \geq 0$ and $\forall h:\mathbb{R}^k\rightarrow\mathbb{R}$, $h(\mathbf{s}_i) \text{ and } z_i \text{ are statistically independent}$.

As discussed in \cite{kleindessner2023efficient, DBLP:conf/iclr/BalunovicRV22}, linear projections (such as PCA or AA) lack the expressive power needed to completely eliminate sensitive information from a dataset. As a result, a sufficiently strong adversary $h$ may still be able to recover part of this information from the projected data, even if the projections are intended to be fair. To address this limitation, and following the approach proposed in \cite{kleindessner2023efficient}, we relax the original requirement and restrict adversaries to linear functions of the form. Moreover, instead of enforcing full statistical independence, we relax this condition to require that the two variables are uncorrelated, meaning their covariance is zero.

If we define $\bar{z} = \frac{1}{n}\sum z_i$ and $\mathbf{z} = \{z_1 - \bar{z}, \dots, z_n - \bar{z}\} \in \mathbb{R}^n$, we can reformulate the fairness requirement as:

\begin{gather*}
    \forall w \in \mathbb{R}^k, b\in \mathbb{R}: \mathbf{w} \mathbf{s}_i^\transpose + b \text{ and } z_i \text{ are uncorrelated} \Leftrightarrow \\
    \forall w \in \mathbb{R}^k, b\in \mathbb{R}: \sum_{i=1}^n (z_i - \bar{z})(\mathbf{w} \mathbf{s}_i^\transpose + b) = 0 \Leftrightarrow \\
    \forall w \in \mathbb{R}^k: \mathbf{z}\mathbf{S}\mathbf{w}^T = 0 \Leftrightarrow \\
    \mathbf{z}\mathbf{S} = 0
\end{gather*}

Thus, we can define FairAA by introducing an additional term in the optimization problem:

$$\argmin_{\mathbf{S}, \mathbf{C}} \norm{\mathbf{X} - \mathbf{SCX}}_F^2 + \lambda \norm{\mathbf{z}\mathbf{S}}_F^2$$

subject to $\norm{\mathbf{s}_n}_2 = 1$, $\mathbf{s}_n \geq 0$, $\norm{\mathbf{c}_k}_2 = 1$, $\mathbf{c}_k \geq 0$, and $\lambda \geq 0$, which acts as a regularization parameter ensuring that the projection does not retain significant information about $z_i$.

In this case, the objective function can also be written in terms of the trace operator as:
$$E = \norm{\mathbf{X} - \mathbf{SCX}}_F^2 + \norm{\mathbf{z}\mathbf{S}}_F^2 = \text{tr} \left( \mathbf{X} \mathbf{X}^\transpose - 2 \mathbf{SCX}\mathbf{X}^\transpose +  \mathbf{SCX}\mathbf{X}^\transpose \mathbf{C}^\transpose \mathbf{S}^\transpose \right) + \lambda \text{tr}\left( \mathbf{z}\mathbf{S}\mathbf{S}^\transpose\mathbf{z}^\transpose \right),$$

and the resulting gradients are:
$$\nabla_{\mathbf{S}} E = 2(\mathbf{SCX} \mathbf{X}^\transpose\mathbf{C}^\transpose -\mathbf{X} \mathbf{X}^\transpose\mathbf{C}^\transpose + \lambda\mathbf{z}^\transpose \mathbf{z} \mathbf{S}),$$
$$\nabla_{\mathbf{C}} E = 2(\mathbf{S}^\transpose\mathbf{S}\mathbf{C}\mathbf{X}\mathbf{X}^\transpose - \mathbf{S}^\transpose\mathbf{X}\mathbf{X}^\transpose).$$

Regarding the complexity order, the $\lambda \mathbf{z}^\transpose \mathbf{z}$ parameter could be precomputed, so the complexity order will be the same as in \cite{morup2012archetypal}.

\section{Extensions}
\label{sec:extensions}

The proposed FairAA framework offers a high degree of flexibility and can be extended to address more complex scenarios beyond its initial formulation \cite{kleindessner2023efficient}. Below, we outline several natural extensions that enhance its expressiveness and broaden its applicability to more realistic settings, including nonlinear structures and multiple sensitive attributes or groups.

\subsection{Kernelizing fair AA}

As shown in \cite{morup2012archetypal}, the updates of the fair AA model indicate that the gradients depend solely on the pairwise relationships, which are represented by the kernel matrix of inner products, $\mathbf{K} = \mathbf{X}\mathbf{X}^\transpose$. Consequently, the fair AA model can be easily extended to kernel-based representations that rely on other pairwise data point relationships (FairKernelAA). This approach can be understood as extracting the principal convex hull in a potentially infinite Hilbert space, analogous to the interpretations of kernelized k-means and PCA.

In this new framework, the gradients are given by the following expressions:
$$\nabla_{\mathbf{S}} E = 2(\mathbf{SCK}\mathbf{C}^\transpose -\mathbf{K}\mathbf{C}^\transpose + \lambda\mathbf{z}^\transpose \mathbf{z} \mathbf{S}),$$
$$\nabla_{\mathbf{C}} E = 2(\mathbf{S}^\transpose\mathbf{S}\mathbf{C}\mathbf{K} - \mathbf{S}^\transpose\mathbf{K}).$$

where $\mathbf{K} = \mathbf{K}(\mathbf{X}) \in \mathbb{R}^{n \times n}$ represents the kernel matrix that encodes the pairwise relationships between the elements of $\mathbf{X}$.

\subsection{Hiding multiple groups}

So far, we have focused on the case where the critical attribute is formed by two distinct groups. However, to extend this to cases where the critical attribute consists of $m$ disjoint groups, we adopt the one-vs-all approach as outlined in \cite{kleindessner2023efficient}.

In this scenario, for each data point $\mathbf{x}_i$, we introduce $m$ one-hot encoded critical attributes $\mathbf{z}^{(1)}, \dots, \mathbf{z}^{(m)}$, where $z_i^{(l)} = 1$ if $\mathbf{x}_i$ belongs to group $l$, and $z_i^{(l)} = 0$ if it does not.

To ensure fairness across multiple groups, we now require that the linear function $h(\mathbf{s}_i)$, representing the projection of $\mathbf{x}_i$, remains uncorrelated with each of the $z_i^{(l)}$ for all $l \in [m]$. This condition implies that $\mathbf{ZS} = 0$, where the $l$-th row of $\mathbf{Z} \in \mathbb{R}^{m \times n}$ is given by $\{z_1^{(l)} - \bar{z}^{(l)}, \dots, z_n^{(l)} - \bar{z}^{(l)}\}$, with $\bar{z}^{(l)}$ being the mean of the $l$-th critical attribute across all data points. As a result, the optimization problem can be solved in a similar manner to the fair AA approach for two groups.

$$\argmin_{\mathbf{S}, \mathbf{C}} \norm{\mathbf{X} - \mathbf{SCX}}_F^2 + \lambda \norm{\mathbf{Z}\mathbf{S}}_F^2$$

subject to $\norm{\mathbf{s}_n}_2 = 1$, $\mathbf{s}_n \geq 0$, $\norm{\mathbf{c}_k}_2 = 1$, $\mathbf{c}_k \geq 0$, and $\lambda \geq 0$, which acts as a regularization parameter ensuring that the projection does not retain significant information about $z_i^{(l)}$.

\subsection{Dealing with multiple critical attributes}

Another extension of the FairAA model arises when we consider multiple critical attributes, such as gender and race, where each attribute could define multiple groups.

Let us assume there are $p$ critical attributes, with the $r$-th attribute defining $m_r$ groups. For each $r \in [p]$, let $\mathbf{Z}_r \in \mathbb{R}^{m_r \times n}$ be the matrix containing the one-hot encoding for the $r$-th attribute groups. By stacking all the $\mathbf{Z}_r$ matrices into a new matrix $\mathbf{Z} \in \mathbb{R}^{(\sum_r m_r) \times n}$, and using this new matrix in the previously defined algorithms, we obtain the fair AA for multiple critical attributes.



\section{Experiments and results} \label{sec:res}

In this section, we present a series of experiments to evaluate the utility and fairness of FairAA. Since fairness has not yet been explored within the AA framework, we compare our proposed method and its variants against the standard AA baseline.

\subsection{Evaluation metrics}

To assess fairness in alignment with our definition, as done in \cite{kleindessner2023efficient}, we evaluate three key metrics in each experiment:

\begin{itemize}
    \item {Explained Variance (EV):} This metric measures how well the original dataset can be reconstructed from the projections onto the archetypes. It reflects the amount of information retained by the model. A \textit{higher explained variance} indicates better reconstruction quality and thus higher utility of the projections. When fairness constraints are introduced, it is expected that the explained variance may decrease slightly due to the added regularization. However, this drop should not be significant; a sharp decline would suggest that fairness is being achieved at the cost of losing too much information, which is undesirable. Therefore, maintaining a reasonably high explained variance is essential to ensure that the model remains useful while being fair.

    \item {Mean Maximum Discrepancy (MMD):} MMD quantifies the distributional difference between the projected data points of different sensitive groups (e.g., gender, race). It is used here to measure how distinct the representations of different classes are. A \textit{lower MMD} value indicates that the projections of different groups are more similar, suggesting that the model is not encoding group-specific biases---hence, lower MMD is desirable for fairness. Conversely, a high MMD implies that the model captures group-related features, potentially leading to biased outcomes.
    
    \item {Linear Separability (LS):} To assess how distinguishable the groups are in the projected space, we train a \textit{logistic regression classifier}---a linear model---to predict the sensitive attribute based on the projections. If the classifier performs well, it implies that the model preserves information about group membership, which is undesirable from a fairness standpoint. Therefore, \textit{lower linear separability} (i.e., lower classification accuracy) indicates greater fairness, as the model's representations do not allow easy separation of the sensitive groups.
\end{itemize}

By jointly analyzing these three metrics, we can evaluate the fairness of each method. An ideal model would achieve high explained variance while maintaining low MMD and low linear separability, ensuring that useful information is preserved without encoding sensitive-group distinctions.

\subsection{Toy datasets}

\begin{figure}[!ht]
  \begin{subfigure}[b]{0.5\linewidth}
    \centering
    \includegraphics[width=\textwidth]{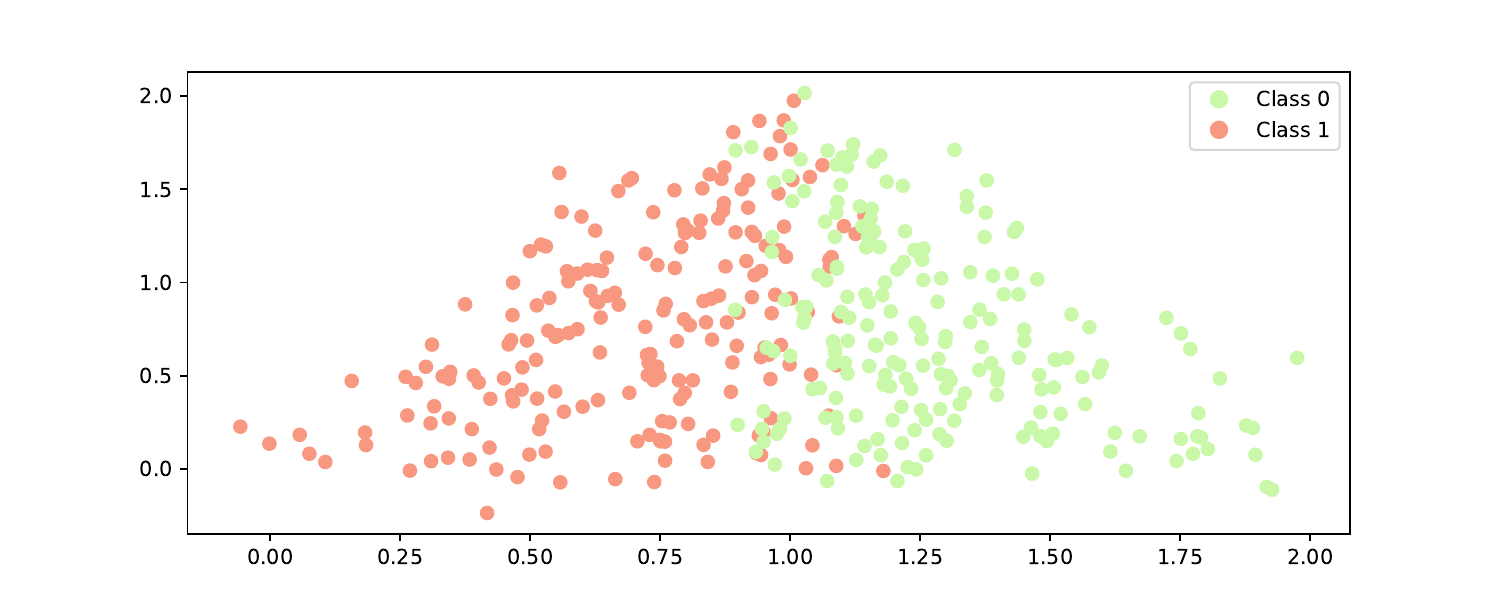}
    \caption{Toy dataset with two classes.}
    \label{fig:dummy-dataset}
  \end{subfigure}
  \hfill
  \begin{subfigure}[b]{0.5\linewidth}
    \centering
    \includegraphics[width=\textwidth]{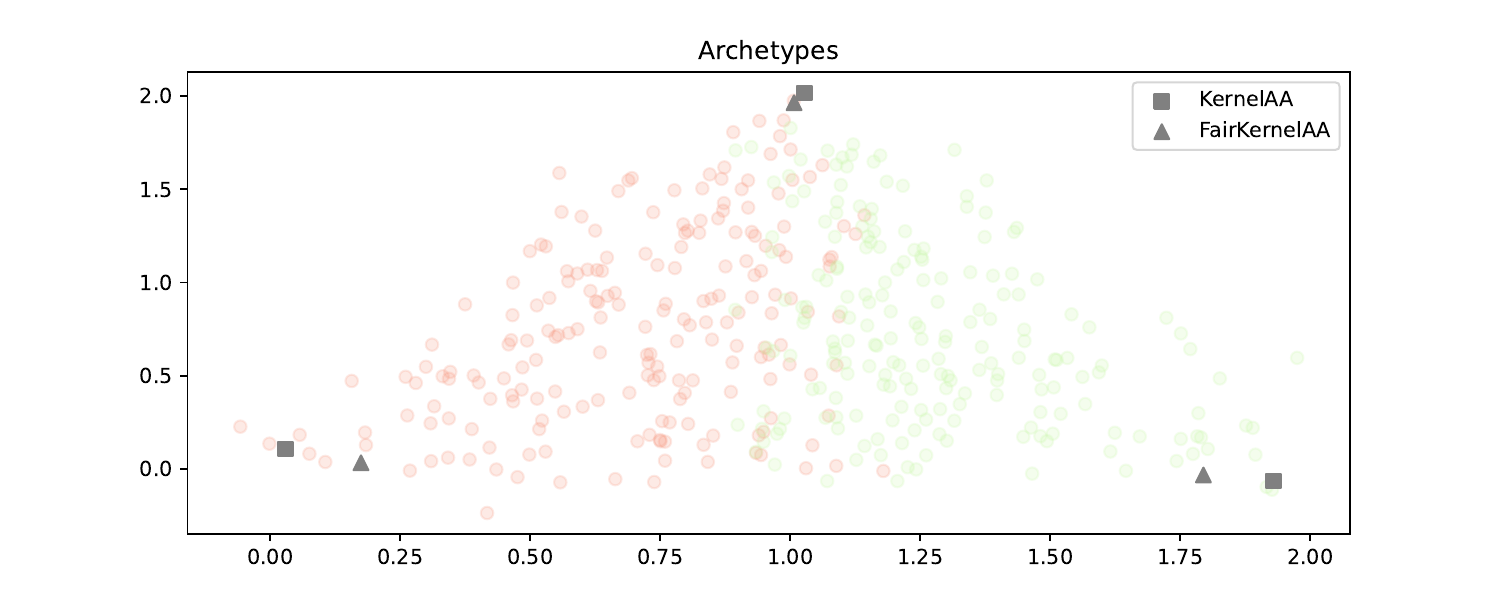}
    \caption{Archetypes discovered by AA and FairAA.}
    \label{fig:dummy-aa}
  \end{subfigure}
\begin{subfigure}[b]{0.5\linewidth}
  \centering
  \includegraphics[width=\textwidth]{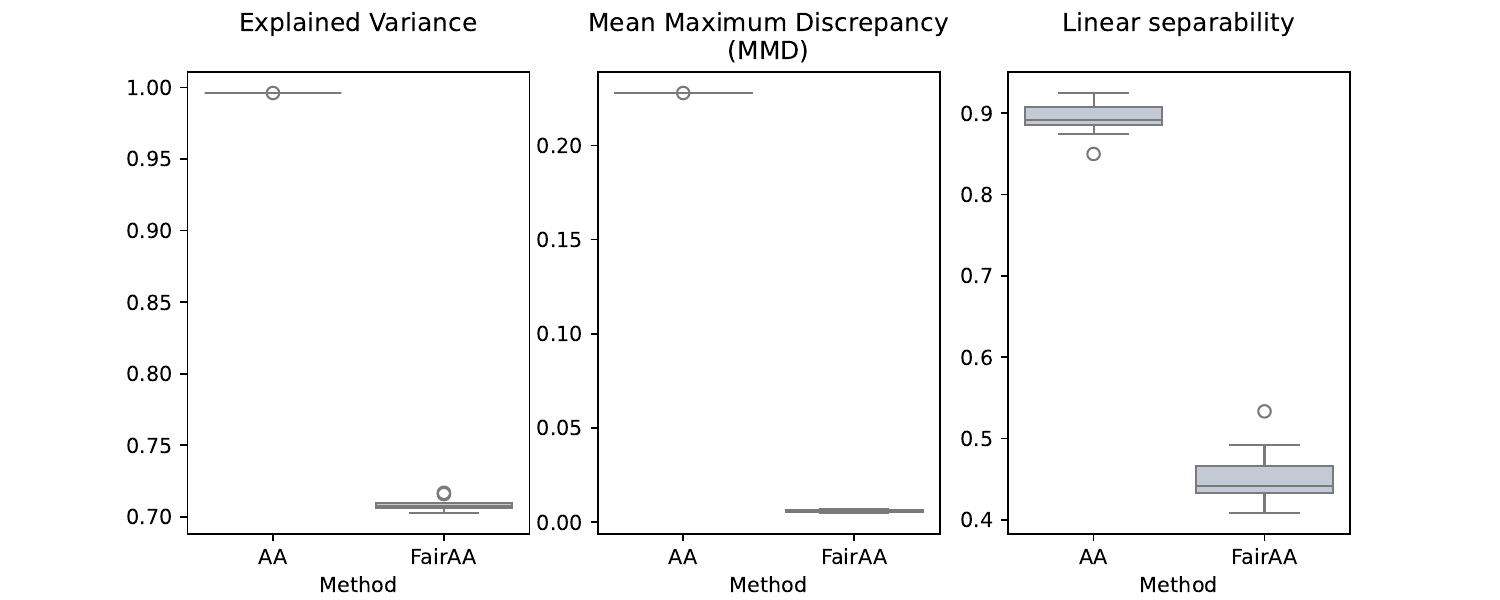}
  \caption{Fairness metrics computed for AA and FairAA.}
  \label{fig:dummy-metrics}
\end{subfigure}
\hfill
\begin{subfigure}[b]{0.5\linewidth}
    \centering
    \includegraphics[width=\textwidth]{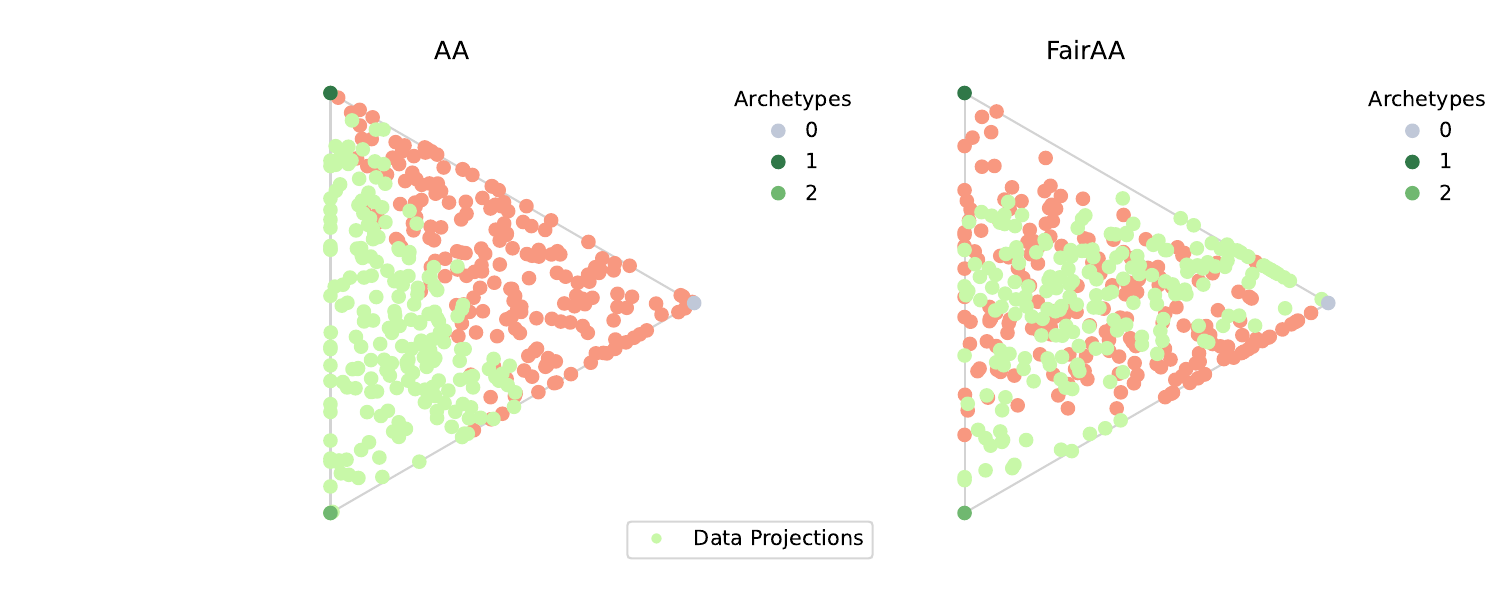}
    \caption{Projection of data onto AA and FairAA spaces.}
    \label{fig:dummy-projection}
  \end{subfigure}
\caption{FairAA vs AA on two-class dataset experiment.}
\label{fig:dummy}
\end{figure}

For the purpose of evaluating FairAA, we produced the dataset shown in Figure~\ref{fig:dummy-dataset} by means of the \texttt{make\_archetypal\_dataset} function from the \texttt{archetypes} Python package \cite{archetypes-package}.

As illustrated in Figures~\ref{fig:dummy-metrics} and~\ref{fig:dummy-projection}, standard AA produces projections where the classes are clearly linearly separable, indicating that group-specific structure is preserved in the representation space. In contrast, FairAA yields projections in which the classes are no longer linearly separable, suggesting that the model effectively mitigates the influence of sensitive attributes. This improvement in fairness occurs despite the archetypes being nearly identical in both models (Figure~\ref{fig:dummy-aa}), confirming that FairAA retains the overall data structure. Quantitatively, FairAA achieves a lower MMD and reduced LS, meaning that group membership is less distinguishable from the representations—two clear indicators of improved fairness. Although the EV is slightly lower than in AA, it remains high, showing that FairAA maintains most of the data variability while achieving a better balance between utility and fairness.

Additionally, in order to explore the nonlinear extension of FairAA, we compared it against standard KernelAA \cite{morup2012archetypal} using a dataset generated via the \texttt{make\_moons} function from the \texttt{scikit-learn} package \cite{scikit-learn-package}. The RBF kernel was chosen over a linear kernel to better model nonlinearities. A total of 8 archetypes were computed to represent the dataset structure.

\begin{figure}[!ht]
    \begin{subfigure}[b]{0.5\linewidth}
        \centering
        \includegraphics[width=\textwidth]{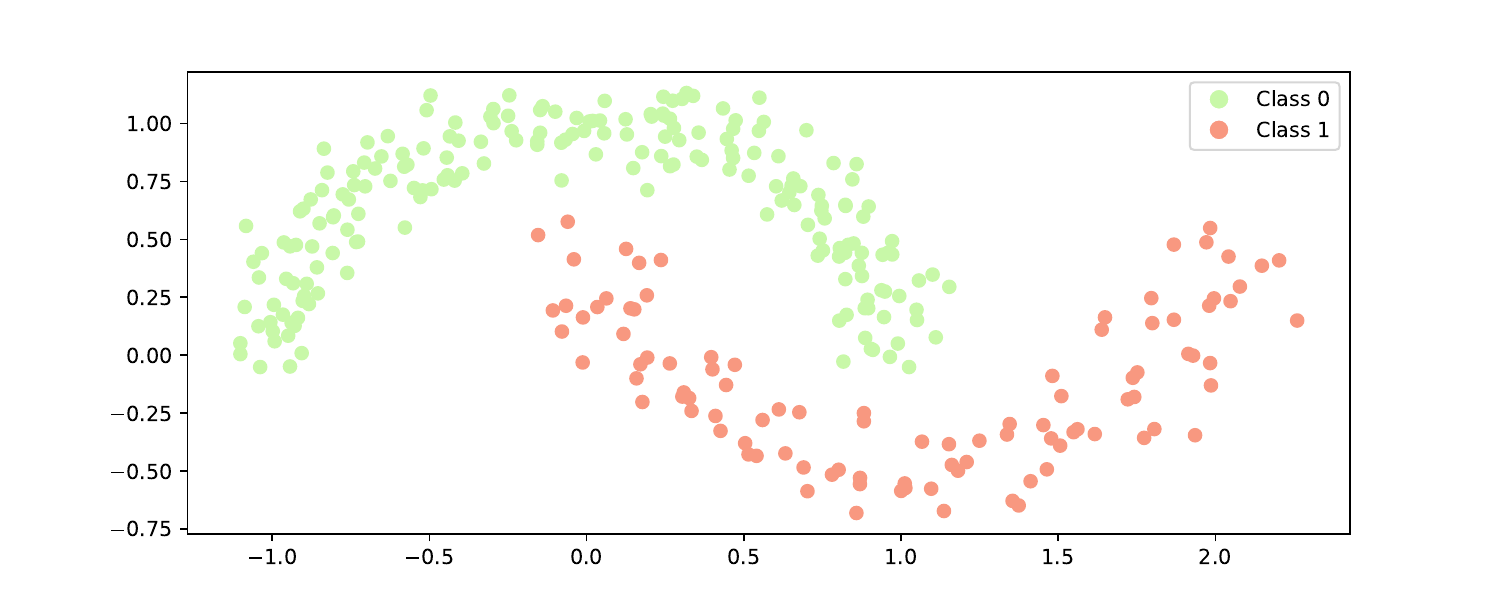}
        \caption{Non-linear, toy dataset with two classes.}
        \label{fig:dummy-kernel-dataset}
    \end{subfigure}
    \hfill
    \begin{subfigure}[b]{0.5\linewidth}
        \centering
        \includegraphics[width=\textwidth]{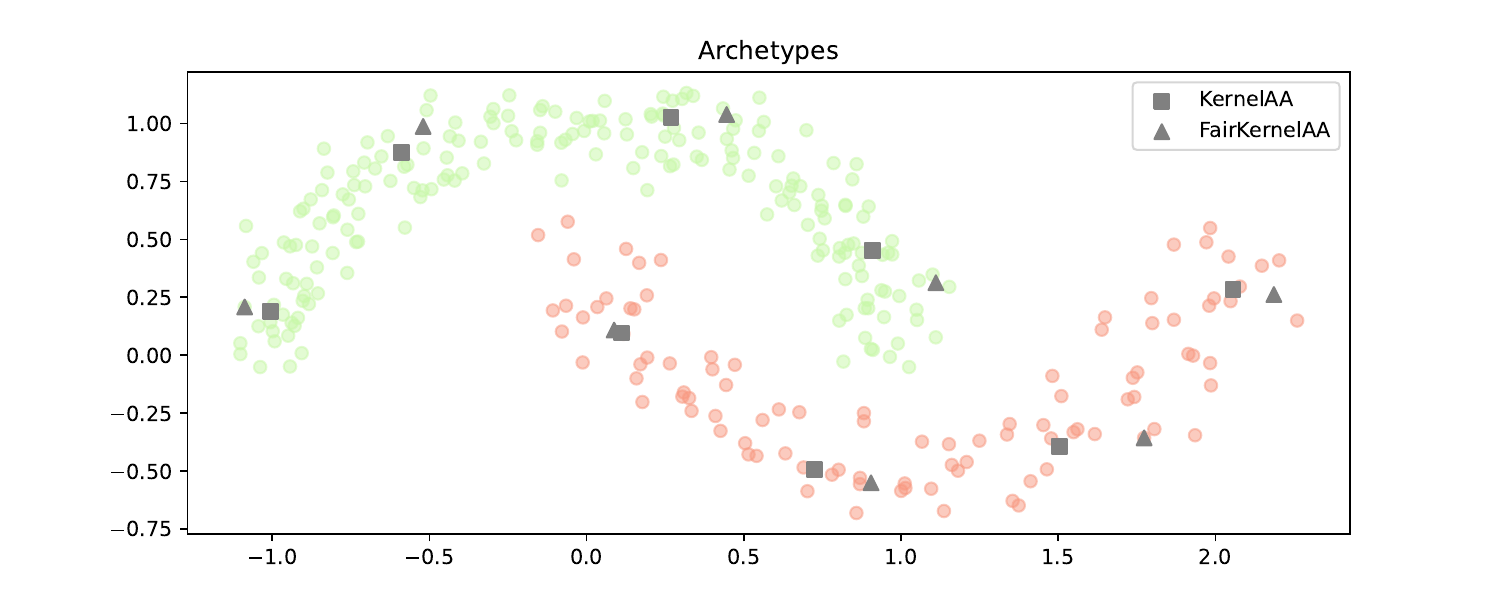}
        \caption{Archetypes discovered by KernelAA and FairKernelAA.}
        \label{fig:dummy-kernel-aa}
    \end{subfigure}

    \begin{subfigure}[b]{0.5\linewidth}
        \centering
        \includegraphics[width=\textwidth]{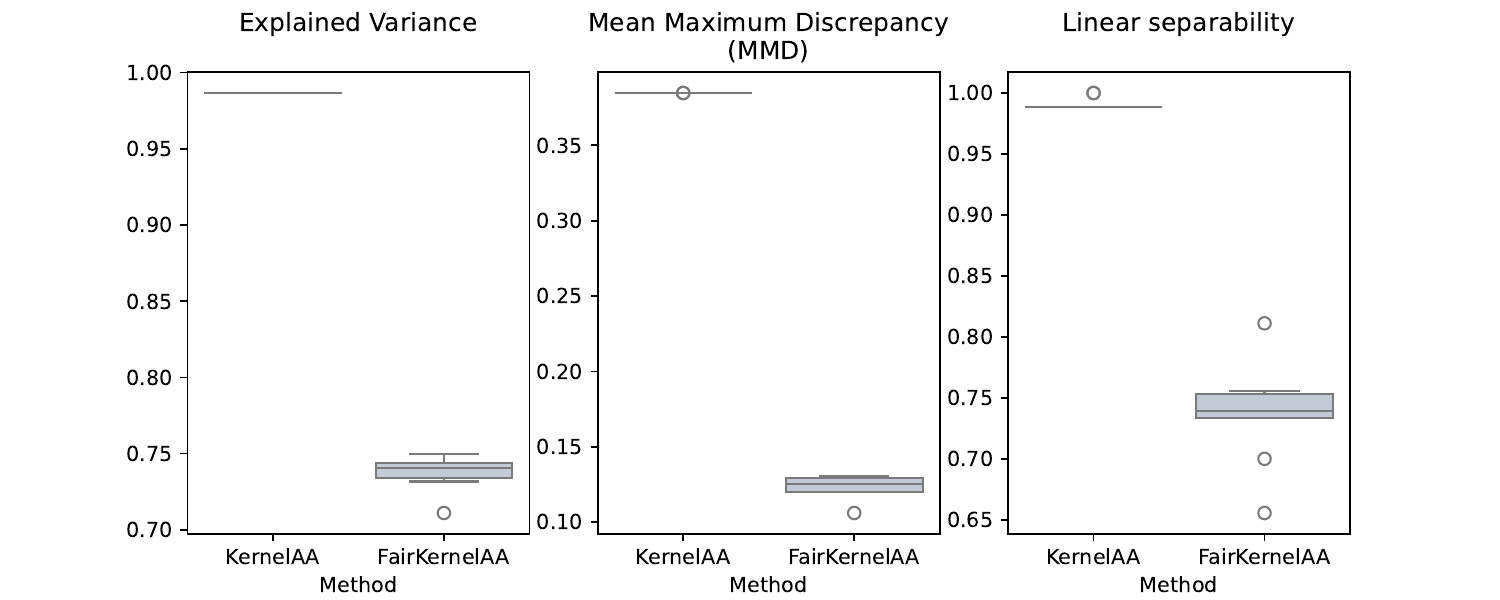}
        \caption{Fairness metrics computed for KernelAA and FairKernelAA.}
        \label{fig:dummy-kernel-metrics}
    \end{subfigure}
    \hfill
    \begin{subfigure}[b]{0.5\linewidth}
        \centering
        \includegraphics[width=\textwidth]{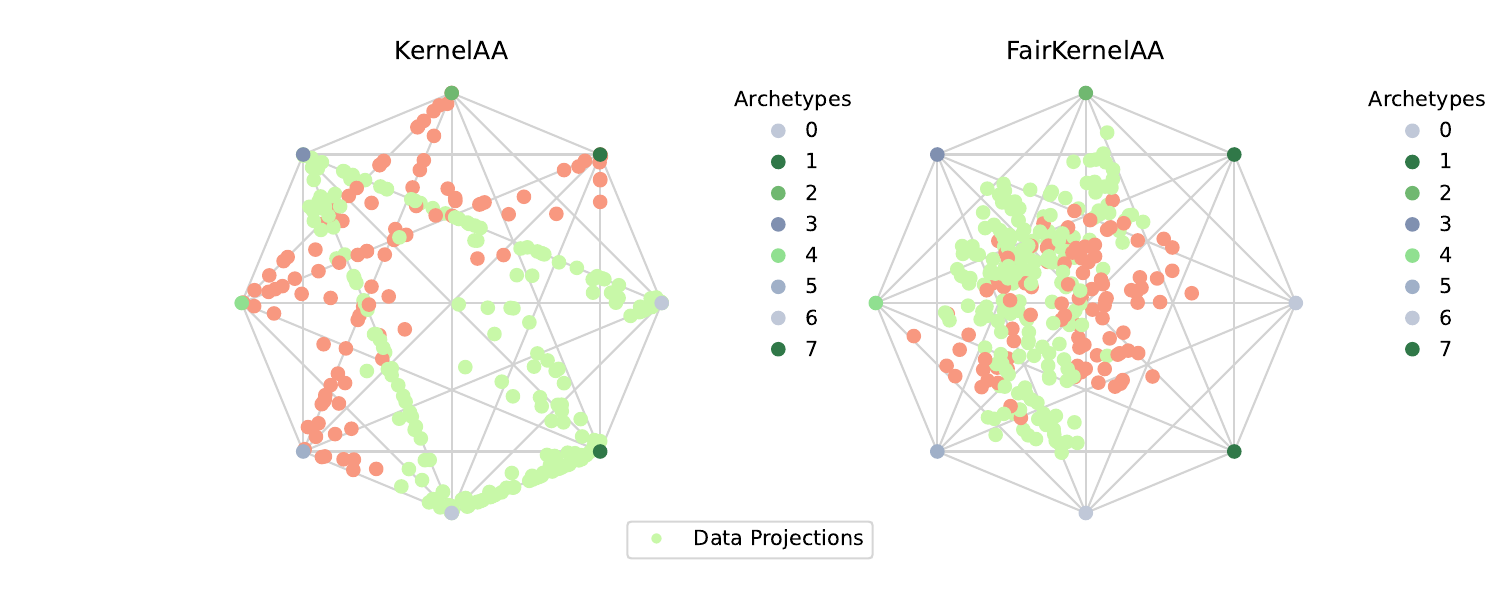}
        \caption{Projection of the data onto KernelAA and FairKernelAA spaces.}
        \label{fig:dummy-kernel-projection}
    \end{subfigure}
    \caption{FairKernelAA vs KernelAA on non-linear dataset experiment.}
    \label{fig:dummy-kernel}
\end{figure}

As shown in Figures~\ref{fig:dummy-kernel-metrics} and~\ref{fig:dummy-kernel-projection}, standard KernelAA produces projections where the classes remain clearly separable, reflecting the preservation of group-specific structure in the representation space. Conversely, FairKernelAA produces projections where class separability diminishes, indicating that the model successfully reduces the influence of sensitive attributes. This enhanced fairness is achieved while maintaining nearly identical archetypes in both models (Figure~\ref{fig:dummy-kernel-aa}), demonstrating that FairKernelAA preserves the overall data structure. Quantitative results show that FairKernelAA attains lower MMD and reduced LS scores, indicating that group membership is less distinguishable in the representations—key measures of fairness improvement. Despite a slight decrease in EV compared to KernelAA, it remains high, confirming that FairKernelAA effectively balances utility and fairness.

Finally, to wrap up the toy experiments, we examined how FairAA performs in scenarios with multiple attributes and groups by applying it alongside standard AA on a four-class dataset. This data was generated using the \texttt{make\_blobs} function from the \texttt{scikit-learn} package \cite{scikit-learn-package}, and 3 archetypes were extracted to capture its underlying structure.

\begin{figure}[!ht]
    \begin{subfigure}[b]{0.5\linewidth}
        \centering
        \includegraphics[width=\textwidth]{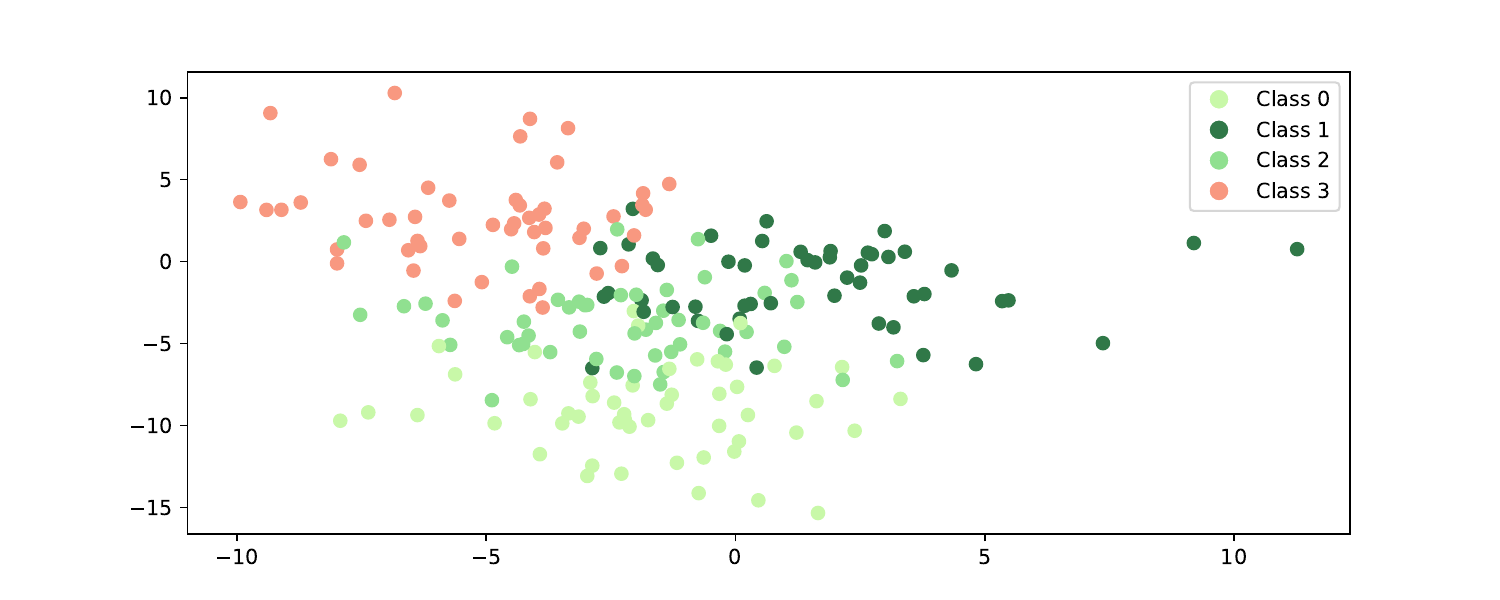}
        \caption{Toy dataset with four classes.}
        \label{fig:dummy-multi-dataset}
    \end{subfigure}
    \hfill
    \begin{subfigure}[b]{0.5\linewidth}
        \centering
        \includegraphics[width=\textwidth]{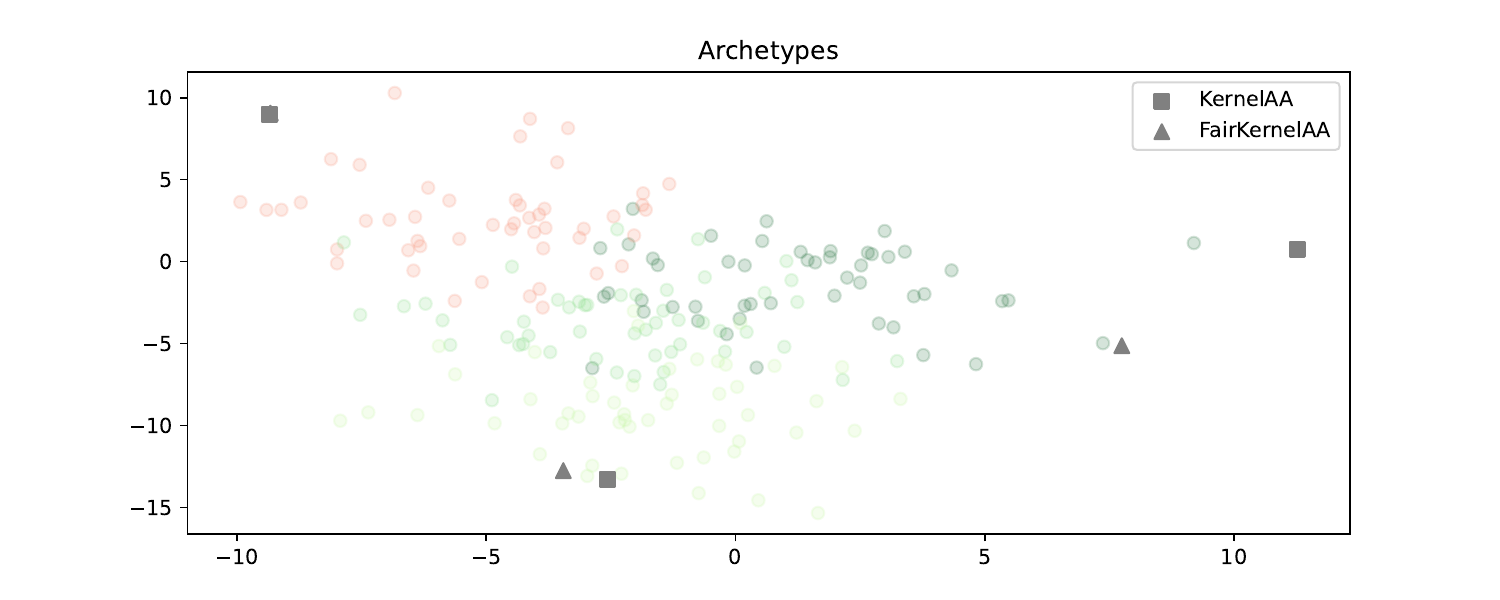}
        \caption{Archetypes discovered by AA and FairAA.}
        \label{fig:dummy-multi-aa}
    \end{subfigure}

    \begin{subfigure}[b]{0.5\linewidth}
        \centering
        \includegraphics[width=\textwidth]{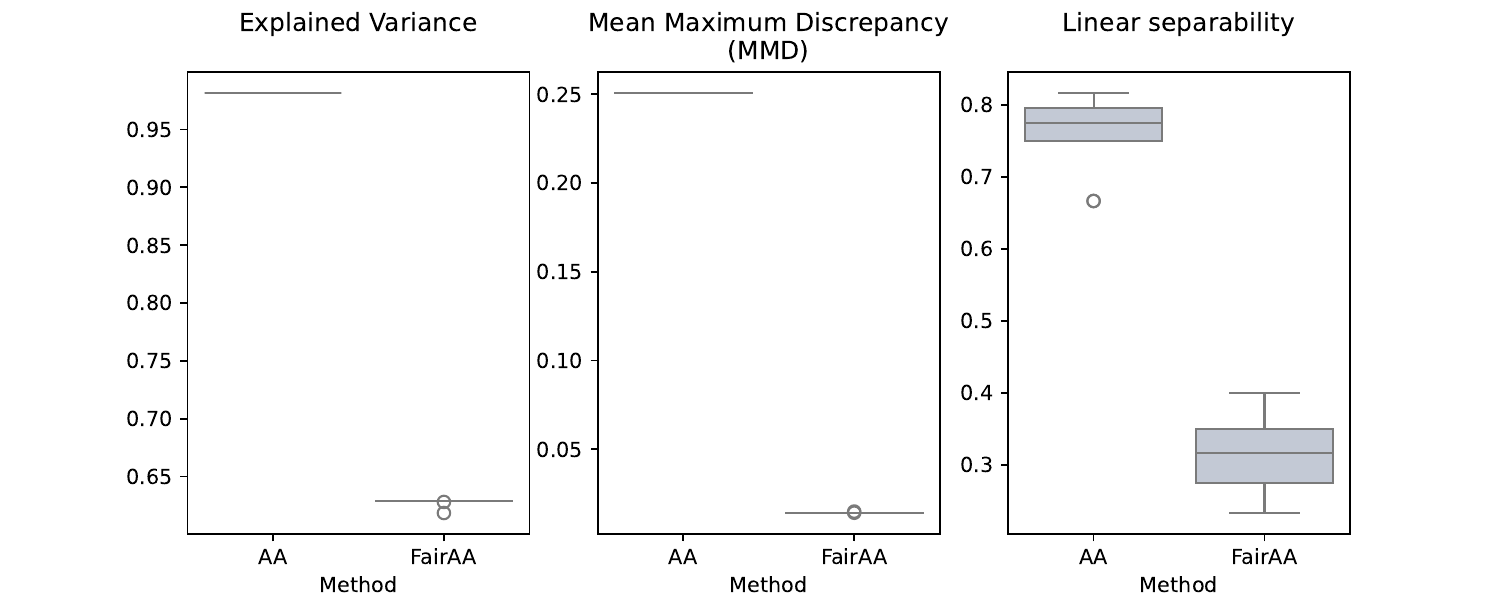}
        \caption{Fairness metrics computed for AA and FairAA.}
        \label{fig:dummy-multi-metrics}
    \end{subfigure}
    \hfill
    \begin{subfigure}[b]{0.5\linewidth}
        \centering
        \includegraphics[width=\textwidth]{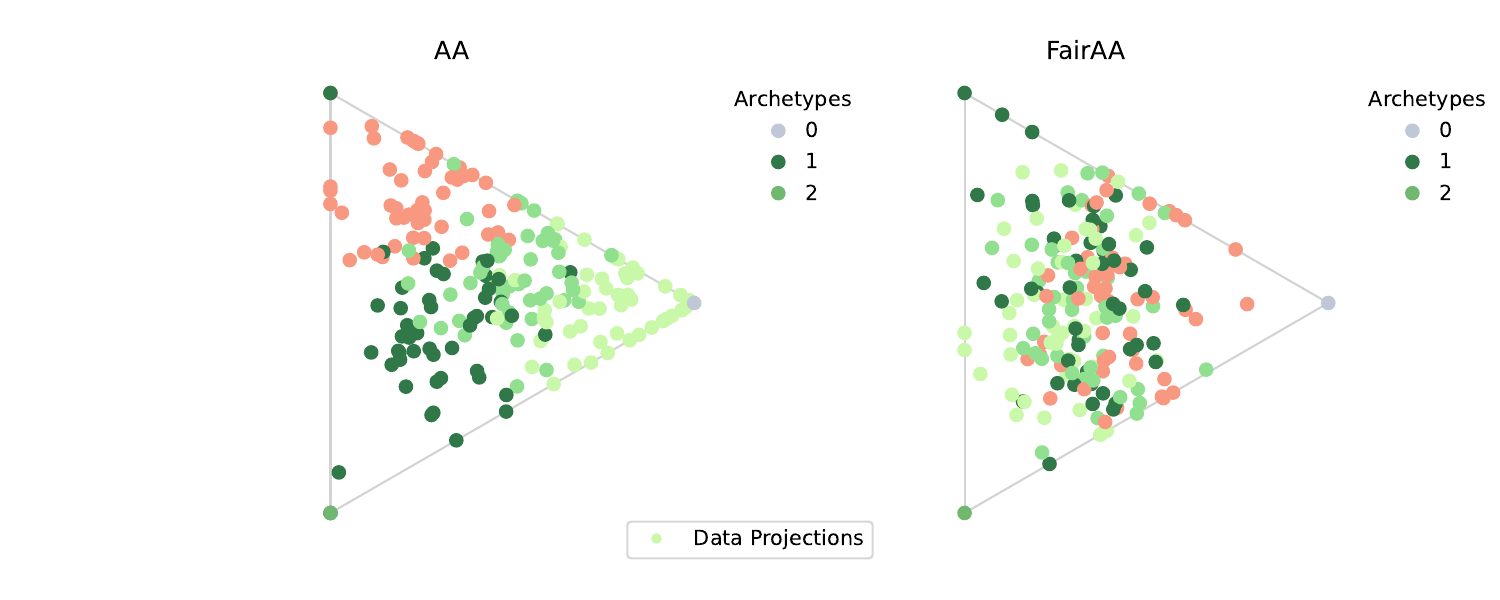}
        \caption{Projection of the data onto AA and FairAA spaces.}
        \label{fig:dummy-multi-projection}
    \end{subfigure}
    \caption{FairAA vs AA on multi-class dataset experiment.}
    \label{fig:dummy-multi}
\end{figure}

As depicted in Figures~\ref{fig:dummy-multi-metrics} and~\ref{fig:dummy-multi-projection}, AA maintains almost clear class separability in its projections, preserving group-specific patterns. In contrast, FairAA reduces this separability, indicating a successful reduction in the impact of sensitive attributes. The archetypes remain largely consistent between models, with two nearly identical and one showing slight variation (Figure~\ref{fig:dummy-multi-aa}), which illustrates that FairAA retains most of the dataset’s structure. From a quantitative standpoint, FairAA yields lower MMD and LS values, signaling improved fairness by making group membership less distinguishable within the representations. Although the EV is somewhat reduced compared to AA, it stays sufficiently high, demonstrating an effective trade-off between utility and fairness.

\subsection{Real Datasets}

After validating the effectiveness of our method using synthetic data, we now evaluate FairAA on real-world datasets. The experiments with dummy data demonstrated that our approach successfully reduces group-specific information in the latent space while preserving the structure and interpretability of AA. However, synthetic examples are by nature controlled and idealized. To assess the practical utility and fairness of FairAA in more realistic scenarios, we turn to real data.

In this section, we use the ANSUR I dataset \cite{gordon19891988}, a widely used anthropometric survey conducted by the U.S. Army, which includes a variety of body measurements collected from male and female personnel. This dataset is particularly relevant for fairness evaluation, as many of its features differ significantly between sexes. If standard AA is applied without considering fairness, the resulting projections may allow for a clear separation of sensitive groups, such as males and females. This separability enables the inference of group membership from the latent representations, which can lead to discriminatory outcomes or biased decisions, even when the sensitive attribute is not explicitly used. Such group-specific encoding compromises fairness by embedding sensitive information into the representation space.
Applying FairAA to this data allows us to test whether our method can mitigate such effects and provide more equitable representations across gender.

Before applying AA, the data was scaled to ensure all features contribute equally to the decomposition. To facilitate visual interpretation of the results, we fixed the number of archetypes to four. The procedure used to evaluate the effectiveness of FairAA is the same as in the previous synthetic examples, relying on the same metrics: explained variance, mean maximum discrepancy, and linear separability.

\begin{figure}[!ht]
  \begin{subfigure}[b]{0.49\linewidth}
    \centering
    \includegraphics[width=1.02\textwidth]{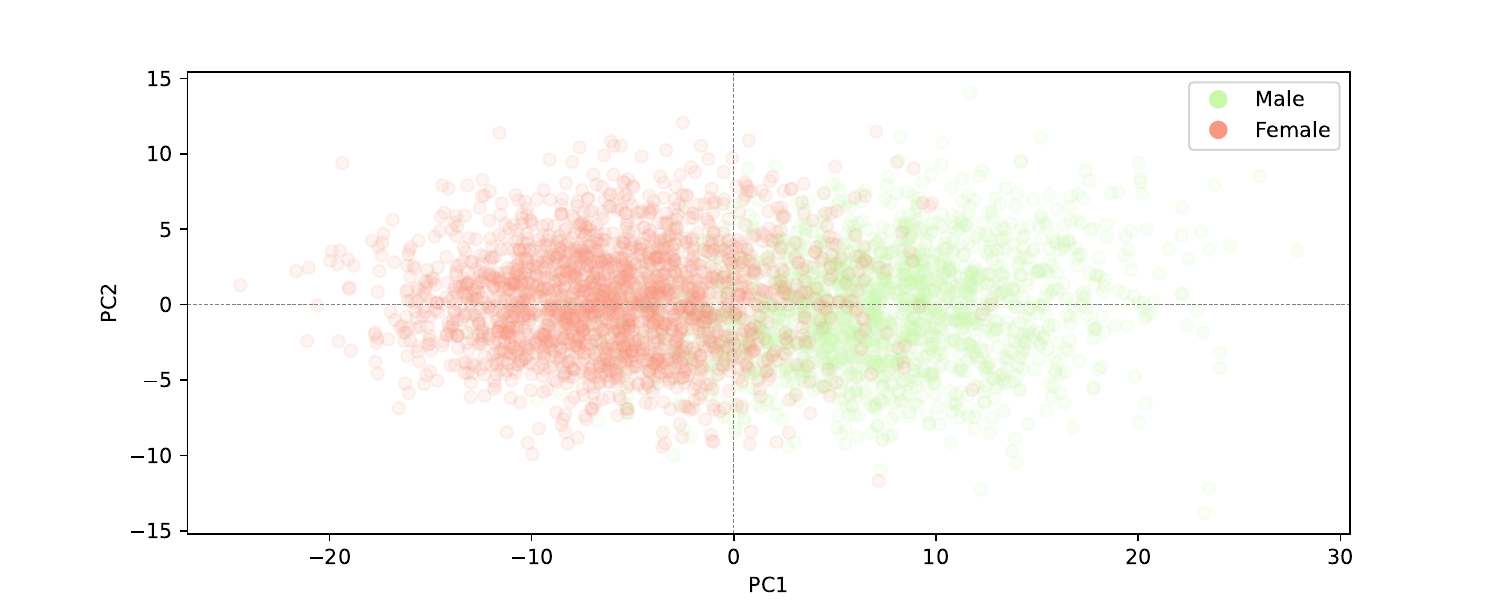}
    \caption{Projection of the ANSUR I dataset onto its first two principal components.}
    \label{fig:antro-dataset}
  \end{subfigure}
  \hfill
  \begin{subfigure}[b]{0.49\linewidth}
    \centering
    \includegraphics[width=1.02\textwidth]{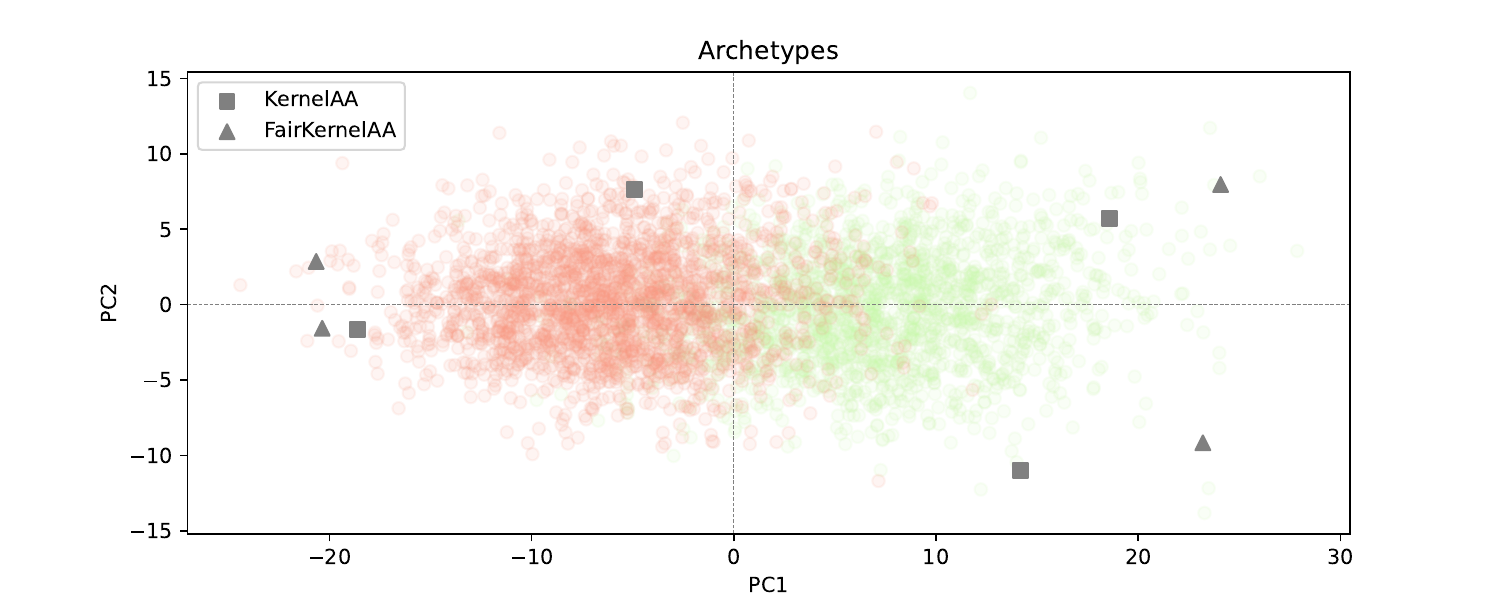}
    \caption{Projection of the archetypes discovered by AA and FairAA onto the first two principal components of the ANSUR I dataset.}
    \label{fig:antro-aa}
  \end{subfigure}
\begin{subfigure}[b]{0.5\linewidth}
  \centering
  \includegraphics[width=\textwidth]{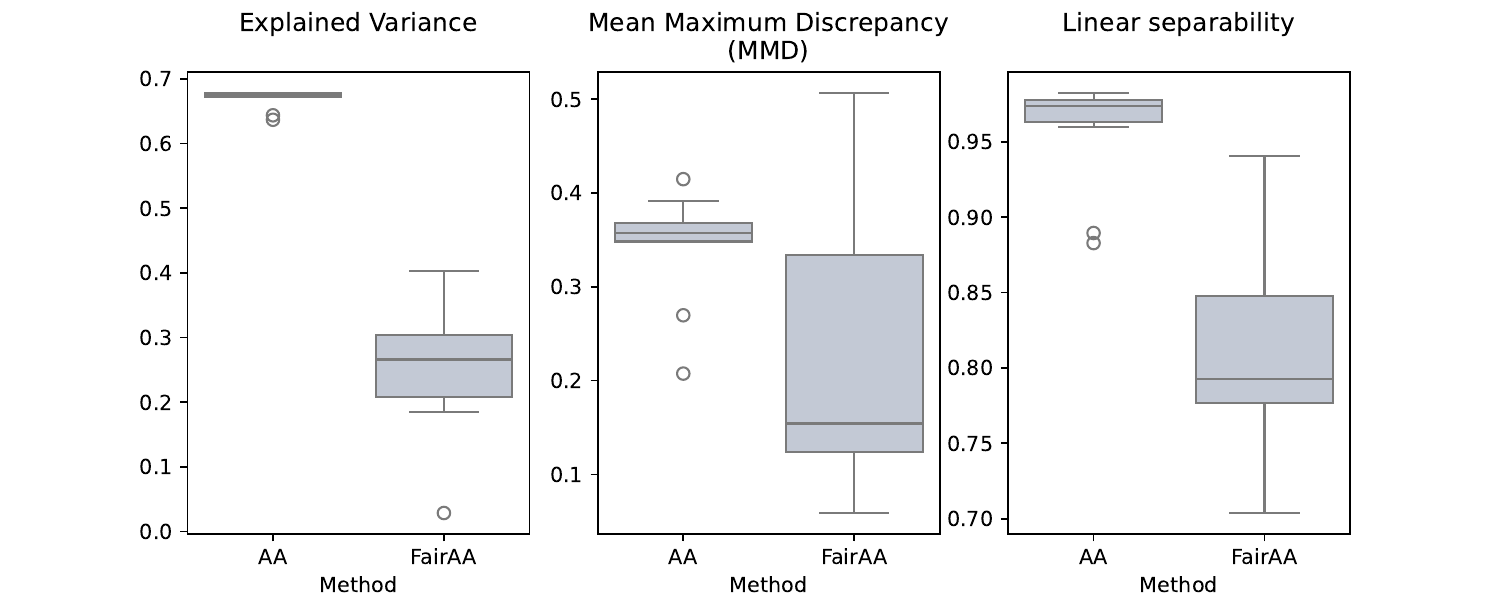}
  \caption{Fairness metrics computed for AA and FairAA.}
  \label{fig:antro-metrics}
\end{subfigure}
\hfill
\begin{subfigure}[b]{0.5\linewidth}
    \centering
    \includegraphics[width=\textwidth]{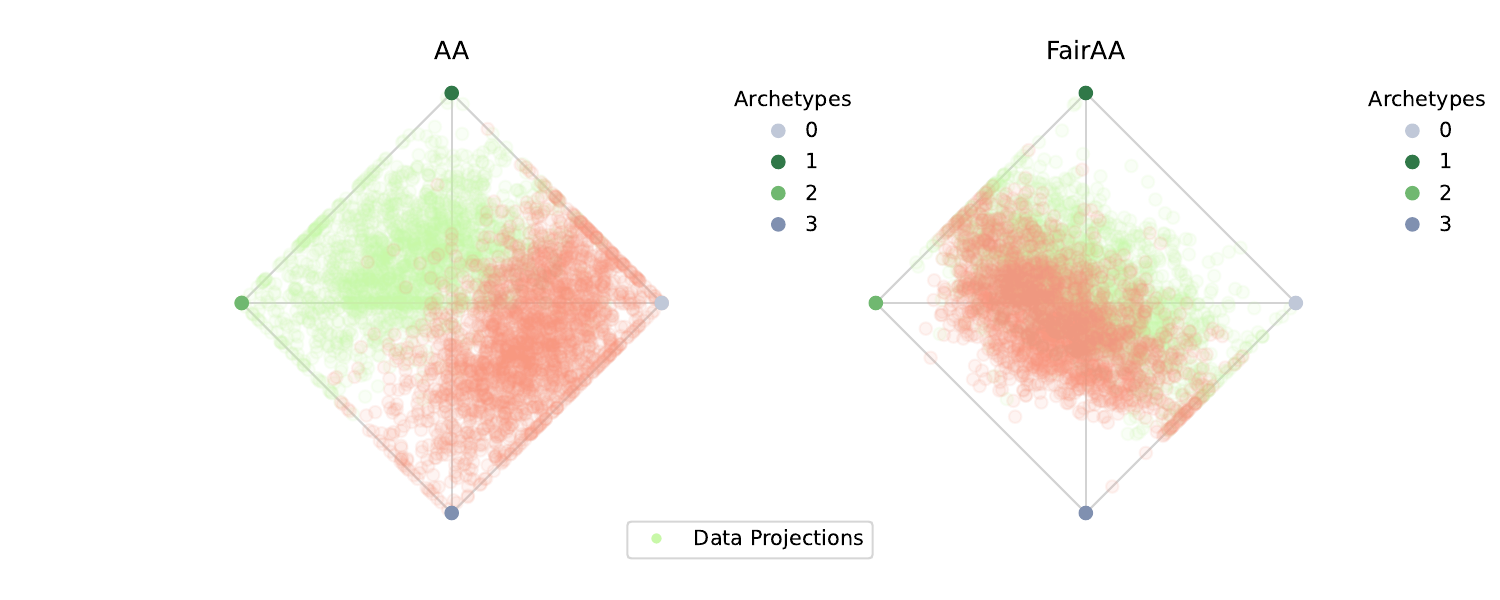}
    \caption{Projection of data onto AA and FairAA spaces.}
    \label{fig:antro-projection}
  \end{subfigure}
\caption{FairAA vs AA on ANSUR I dataset.}
\label{fig:antro}
\end{figure}

The results obtained on the ANSUR I dataset, shown in Figures~\ref{fig:antro-metrics} and \ref{fig:antro-projection}, align with those from the synthetic experiments. Standard AA generates projections in which male and female individuals are clearly linearly separable, indicating that group-specific structure is retained in the latent space. In contrast, FairAA produces projections with much lower separability and reduced group discrepancy, as indicated by the Mean Maximum Discrepancy. Although the explained variance is slightly lower than in the standard model, it remains high enough to preserve the essential structure of the data. These results confirm that FairAA improves fairness while maintaining a high level of utility, also when applied to real-world data.

\section{Conclusions} \label{sec:con}

Fairness in machine learning has become a crucial concern in recent years, as models are increasingly deployed in sensitive applications affecting individuals and society at large. Algorithms like FairAA and FairKernelAA that explicitly address fairness while preserving the utility of the archetypes obtained are essential to ensure equitable and trustworthy outcomes.

Our experiments on diverse synthetic datasets demonstrate that FairAA and its nonlinear extension, FairKernelAA, effectively reduce the distinguishability of sensitive group membership, as measured by lower MMD and LS scores, without substantially sacrificing explained variance. This balance confirms the models’ ability to mitigate bias while maintaining meaningful data structure. The consistent preservation of archetypes across experiments highlights the robustness of the approach in various scenarios, including linear, nonlinear, and multi-group settings.

We further validated the effectiveness of FairAA on the real-world ANSUR I dataset, which contains anthropometric measurements from male and female individuals. The results on this dataset confirm the trends observed in synthetic experiments. Standard AA produces projections where the sensitive attribute (sex) can be easily inferred, raising fairness concerns due to the potential for discriminatory downstream use. In contrast, FairAA significantly reduces the separability between groups while preserving high explained variance and maintaining the overall structure of the archetypes. This demonstrates that FairAA is not only effective in controlled settings but also robust and applicable to real-world data, where fairness is critical to avoid unintended harm.

Future work could extend the FairAA framework to incorporate broader notions of fairness beyond merely equalizing the means of group-specific projections. For example, subsequent research might aim to balance reconstruction errors across groups to prevent any group from experiencing disproportionately higher errors. Moreover, exploring techniques that align both the means and covariances of the group representations could promote a more comprehensive form of statistical parity.

Given the societal impact of biased algorithms, advancing fairness-aware techniques such as FairAA is a timely and necessary step towards more responsible and ethical machine learning.

\section*{Acknowledgement}
This work was partially supported by the Spanish Ministry of
Science and Innovation PID2022-141699NB-I00 and PID2020-
118763GA-I00 and Generalitat Valenciana
CIPROM/2023/66.

\bibliographystyle{abbrv}
\bibliography{references}

\end{document}